%% file: pre-print.tex
\documentclass{article}

\usepackage{arxiv}

\usepackage[utf8]{inputenc} 
\usepackage[T1]{fontenc}    
\usepackage{hyperref}       
\usepackage{url}            
\usepackage{booktabs}       
\usepackage{amsfonts}       
\usepackage{nicefrac}       
\usepackage{microtype}      
\usepackage{lipsum}		
\usepackage{graphicx}
\usepackage{natbib}
\usepackage{doi}

\title{State of NLP in Kenya: A Survey}

\author{       
	Cynthia Jayne Amol  \\
	Maseno University \\
	\And
	{Everlyn Asiko Chimoto} \\
        University of Cape Town \\
        African Institute for Mathematical Sciences - AIMS\\
	\AND
	{Rose Delilah Gesicho} \\
	United States International University-Africa\\
        Zindi \\
       \And
	{Antony M. Gitau}  \\
	African Institute for Mathematical Sciences - AIMS\\
        \And
	{Naome A. Etori  } \\
	  Jomo Kenyatta University of Agriculture and Technology - JKUAT  \\
        The University of Minnesota, Twin Cities \\
	\And
	{Caringtone Kinyanjui} \\
	University of Manchester \\
        Songhai\\
	\And
	{Steven Ndung’u } \\
	University of Groningen \\
        Stellenboch University\\
              \And
	{ Lawrence Moruye } \\
	Jumia Group \\
        \And
	{Samson Otieno Ooko } \\
	Adventist University of Africa  \\
        \And
	{ Kavengi Kitonga  } \\
	University of Nairobi  \\
        \And
	{Brian Muhia } \\
	Independent Researcher  \\
        \And
	{Catherine Gitau} \\
	African Institute for Mathematical Sciences - AIMS  \\
        \And
	{Antony Ndolo}   \\
	Deep Learning Indaba \\
        Karadeniz Technical University\\
        \And
        {Lilian D. A. Wanzare } \\
	Maseno University \\
        \And
	{Albert Njoroge Kahira} \\
	Datawise Africa \\	
        \And
	{Ronald Tombe } \\
	Kisii University\\
        Future Africa - University of Pretoria\\
}

\begin{document}
\maketitle

\begin{abstract}
Kenya, known for its linguistic diversity, faces unique challenges and promising opportunities in advancing Natural Language Processing (NLP) technologies, particularly for its underrepresented indigenous languages. This survey provides a detailed assessment of the current state of NLP in Kenya, emphasizing ongoing efforts in dataset creation, machine translation, sentiment analysis, and speech recognition for local dialects such as Kiswahili, Dholuo, Kikuyu, and Luhya. Despite these advancements, the development of NLP in Kenya remains constrained by limited resources and tools, resulting in the underrepresentation of most indigenous languages in digital spaces. This paper uncovers significant gaps by critically evaluating the available datasets and existing NLP models, most notably the need for large-scale language models and the insufficient digital representation of Indigenous languages. We also analyze key NLP applications—machine translation, information retrieval, and sentiment analysis—examining how they are tailored to address local linguistic needs. Furthermore, the paper explores the governance, policies, and regulations shaping the future of AI and NLP in Kenya and proposes a strategic roadmap to guide future research and development efforts. Our goal is to provide a foundation for accelerating the growth of NLP technologies that meet Kenya's diverse linguistic demands.
\end{abstract}


\input{introduction}

\input{background}
\input{Methodology}
\input{datasets}

\input{applications}
\input{governance}

\input{indigenous}
\input{discussion}

\bibliographystyle{ACM-Reference-Format}
\bibliography{refs}

\end{document}

%% file: introduction.tex
\section{Introduction}


The last few years have seen an explosion in the number of Natural Language Processing(NLP) applications and tools. Tools such as chatGPT are now ubiquitous in many aspects of our lives.  NLP subfield of artificial intelligence (AI) that seeks to enable machines to understand, interpret, and generate human language. Although there have been significant advances in the field of NLP, most of the research today focuses on less than 1\% of the world's languages with a particular focus on ten or so languages, including English \cite{joshi2020state,bender2019benderrule}. One of the main challenges of NLP, especially in low-resource settings, is the availability of requisite datasets to train models in low-resource languages. This is a big issue regarding the progress of NLP for low-resource languages \cite{siminyu2021ai4d,ruder20194}.  With the increasing number of research interests in NLP globally, multiple efforts have emerged in Africa to create both NLP tools and datasets that power NLP technology, reflecting the continent's drive toward AI democratization while addressing concerns about digital colonization \cite{etori2024double} and the need for locally grounded solutions. 

Africa is one of the most linguistically diverse regions in the world, with an estimated 3,000 languages spoken across the continent. However, efforts to forge cohesive nation-states from multiple ethnic groups have often resulted in the marginalization of several languages. Priority has typically been given to high-resource languages such as English and French, alongside regional or national languages like Kiswahili and Afrikaans \cite{adebara2024towards,yandecolonizing}.  While these dominant languages are widely spoken, millions of Africans communicate in at least three languages, including one or two national and indigenous languages. Unfortunately, many of these indigenous languages remain underrepresented in the digital space despite their prevalence. 

In Kenya, there has been limited progress in mapping the country's languages and developing digital tools to support them, particularly for the indigenous languages spoken by millions. This highlights the need for a comprehensive assessment of the current state of NLP technologies in Kenya, considering its diverse geographical and linguistic contexts. Many of these languages remain underrepresented in digital spaces, underscoring the urgency for targeted efforts to bridge this gap and accurately reflect NLP advancements in the region.




\begin{table}
\centering
\footnotesize
\caption{Main language groups  in Kenya with Number of speakers }
\begin{tabular}{p{2.5cm}p{3.5cm}p{4.5cm}}
\hline
\textbf{Bantu} & \textbf{Nilotic} & \textbf{Cushitic} \\
\hline
Kikuyu 8.1 million & & \\
Kamba 4.7 million & Dholuo 5.0 million & \\
Luhya 10 million & Kalenjin languages 4.6 million & Oromo (over 48 million incl. Ethiopia) \\
Gusii 2.7 million & & Borana, 3.4 million speakers in 2010 \\
Meru 2.0 million & Maasai 1.2 million & Orma, 659,000 speakers in 2015 \\
Mijikenda/Giriama & Turkana 1.0 million & Somali 2.8 million (22 million incl. Ethiopia and Somalia) \\
ca. 1 million & & \\
\hline
\end{tabular}
\label{tab:main_languages}
\end{table}

Kenya is made up of multilingual communities. First, it is important to distinguish the difference between tribes and languages. A widely cited but controversial number is the 42 tribes.   One would expect, therefore, that Kenya would have about 42 languages based on the tribes. However, the total number of tribes in Kenya is not well researched, neither are there clear distinctions on what is a tribe. According to ethnology \cite{gordon2005ethnologue}, there are about 68 spoken languages in Kenya today.  Most of the research attempts to study language groups instead of individual languages.
Kenyan languages are generally fall into 3 groups; Bantu, Nilotic and Cushitic. Table \ref{tab:main_languages} shows the most widely spoken languages in their language groups.  With a population of 50 million and growing, doubled by a recent emergence of language and culture appreciation, the roadmap in this paper provides a path towards digital representation for this population.

Some efforts have started in Kenya to collect, clean and organise language datasets in several indigenous languages. These datasets have already been used to develop NLP tools in several of Kenyans languages\footnote{We highlight such efforts in subsequent sections of the paper}. With this backdrop, this paper surveys on the state of NLP in Kenya, aiming to highlight ongoing work and bring out the challenges and opportunities in the field.  As these efforts continue, it is  important to record and map them and create a road map for the future. This survey is an initial attempt at this.

To the best of our knowledge, this  survey is the first  attempt to consolidate all efforts by researchers, practitioners and linguists and provide a reference point for NLP research in Kenya. It is also  the first major attempt to map Kenyan languages from an NLP perspective and provide an initial perspective of a prospective research road map for NLP research in Kenya. While the field of NLP is very broad and keeps growing, we aim to cover as many subtopics are possible. 

The remainder of this paper is organised as follows; first, we look at similar surveys and works in Section \ref{secRelatedwork}. In Section \ref{secDataset} , we survey all datasets on Kenyan languages. In section \ref{secapplication} we survey NLP applications developed in Kenya or using kenyan languages. We touch on categories of applications such as machine translations, information retrieval, and text classification. In section \ref{secgov} we look at governance, policies and regulations related to AI and NLP specifically. We highlight different efforts by governments and communities to draft AI policies and regulations around the technology. Finally in section \ref{secDiscussion} we discuss a possible roadmap for NLP research in Kenya.

%% file: background.tex
\section{Related Work }
\label{secRelatedwork}



The systematic review by \citet{chesire2024current} highlights an expanding corpus of research in dataset creation, machine translation, and the development of multilingual pretrained models tailored for African languages. Prominent initiatives include the development of datasets such as MASAKHANEWS, which facilitates news classification in 16 African languages \cite{adelani2023masakhanews}, and AfriSenti, designed for sentiment analysis across 14 African languages \cite{muhammad2023afrisenti}. Advancements in machine translation are evident, with models like AfriByT5 \cite{adelani2022few}  and MMTAfrica \cite{emezue2022mmtafrica} catering to the linguistic diversity of the continent. Nonetheless, the authors point out that the African NLP landscape still confronts considerable hurdles, chiefly the dearth of extensive datasets and a limited number of NLP researchers within the region.

\citet{adebara2022towards} assess the Afrocentric NLP for the African languages. In their paper, they explore the linguistic diversity of African languages and their distinct challenges for NLP, including tonal systems, vowel harmony, and serial verb constructions. It emphasizes that these linguistic traits are underrepresented in most mainstream NLP languages, underscoring the necessity for specialized methodologies tailored to African languages. Furthermore, the paper sheds light on various sociopolitical factors, such as national language policies that prioritize foreign languages, low literacy rates in indigenous languages with individuals often only literate in foreign tongues, and the absence of standardized orthographies or the presence of inconsistent spelling norms, all of which add complexity to text processing and analysis. These issues profoundly affect the advancement of NLP for African languages.

There have been attempts to survey the state of NLP in Africa from regional and country-specific perspectives. 
From a regional perspective, the study by \cite{mussandi2024nlp} discusses some of the opportunities and challenges of building technologies for languages of African origin in their survey on NLP tools for African languages. It features corpora and task-specific language models developed by previous studies for these languages, exposing the scarcity of these NLP tools that contribute to African languages' ‘technological delay’. AI4D - Artificial Intelligence for Development initiative \cite{siminyu2021ai4d} noted the technological gap created by lack of datasets in African languages. This initiative, aimed at creating datasets for African languages through crowd-sourcing data, reported the setbacks of creating data resources for these languages.
\cite{hedderich2020survey} review NLP approaches for low-resource languages touching on issues that affect African languages, where most are low-resource.


From country specific perspectives, \cite{azunre2021nlp} in their study on the state of NLP in Ghana highlight some of the efforts of NLP Ghana in developing data sources and language tools for Ghaninan languages that are considered low-resource. Some of their contributions such as building annotated datasets, embedding models and translators and  for most-widely spoken languages in Ghana are aimed at increasing availability of language resources for these languages. 
\cite{marivate2020african} highlights the South African NLP landscape, zooming into the existing disparity when it comes to availability of language content in the various South African languages on Wikipedia. The study also looks at past experiences of language processing for South African languages and paints a way forward for African NLP through community building. 


.

%% file: Methodology.tex
\section{Methodology}
\label{secMethodology}
This research follows a multi-step approach encompassing data collection, analysis of existing models, and evaluation of available resources for Kenyan languages to assess the current state of NLP technologies in Kenya. Our methods involved reviewing locally and internationally published research and datasets, as long as they incorporated Kenyan languages, ensuring a comprehensive analysis.

We gathered relevant data by extensively reviewing publicly available datasets, research papers from Google Scholar, ACL, etc., and NLP tools. We focused on Kenyan languages such as Swahili, Dholuo, and Luhya. Our survey included datasets for text classification, machine translation, sentiment analysis, and question answering.  

Lastly, we compared the available datasets and models with Kenya's languages and identified significant gaps in data availability and technological adaptation. Our comparison followed a structured approach focused on five key areas: \textbf{ language coverage}, \textbf{ NLP tasks}, \textbf{NLP tools },  \textbf{  NLP application tasks}, and  \textbf{ resource availability}. We assessed the representation of Kenyan languages in available datasets. The datasets were analyzed using the application of NLP tasks. We examined the practical application of these tools in functions such as translation and speech recognition. Lastly, we compared the availability of computational resources, including annotated datasets and pre-trained models, across Kenyan languages, identifying critical gaps.

%% file: datasets.tex
\section{Datasets}
\label{secDataset}


\begin{table*}[ht]
\caption{Published Kenyan Datasets for Text and Speech} 
\centering
\footnotesize
\begin{tabular}{p{0.3cm} p{5cm} p{5cm} p{1.5cm} p{3cm}} 
\toprule
\# & \textbf{Dataset Name}     & \textbf{Available}    & \textbf{Venue}      & \textbf{Language}            \\ 
\midrule

1  & Kencorpus: Kenyan Languages  & ML/NLP Dataset~\cite{wanjawa2022kencorpus}  & Maseno & Swahili|Luo| Luhya  \\

2  & Building African Voices   & Speech synthesis Dataset~\cite{ogayo2022building} & Interspeech & Suba|Luo|Swahili \\ 

3 & CMU Wilderness  & Multilingual Speech~\cite{CMUWilderness_8683536} &  IEEE     &  Oromo|Somali|Sabaot \\ 

4 & Bible TTS  &  Speech Dataset~\cite{meyer2022bibletts}& Interspeech & Kikuyu|Luo\\ 

5  & Building Low Resource Datasets  & Text and Speech Dataset~\cite{babirye2022building}& AfricaNLP  & Kiswahili \\ 

6 &  XTREME-S &  Cross-lingual Dataset~\cite{conneau2022xtreme} & Interspeech &  Luo|Somali|Swahili|etc\\                   

7 & The African Story Book(ASb) & multilingual African Stories~\cite{stranger2017african}  & SA & Kisii|Kikuyu|Swahili|etc \\ 

8  & Common Voice   & Multilingual Speech Dataset~\cite{ardila2019common} & Mozilla  & Luo|Swahili|Somali|etc \\

9 & IARPA Babel  & Telephone speech~\cite{AB2_HSAU9N_2022} & \ Appen  & Dholuo \\ 

10  &  Kiswahili TTS dataset & TTS Dataset~\cite{Rono_2021} & Mendeley & Kiswahili \\  

11  &  Swahili audio mini-kit & Audio Dataset & TWB & Kiswahili \\ 

12  &  Bloom-lm & Audio Dataset~\cite{leong2022bloom} & SIL-AI & Kiswahili|Kamba|Luo \\ 

13  &  AfriSpeech-200 & ASR Dataset~\cite{olatunji2023afrispeech} &TACL& Kenyan-English \\

14  &  AfroDigits & Speech Dataset~\cite{chinenye2023afrodigits} &AfricaNLP & Oromo\\

15  &  1000 African Voices & Speech synthesis Dataset~\cite{ogun20241000} & Interspeech & Kenyan-eng\\

16  & Helsinki Corpus & Text Dataset~\cite{hurskainen2004helsinki} & Kielipankki & Kiswahili \\ 

17  &  MasakhaNEWS & Text Classification Dataset~\cite{adelani2023masakhanews}  &  IJCNLP & Kiswahili \\

18  & AFRIHG & Text Summarization Dataset~\cite{ogunremiafrihg}  &  AfricaNLP & kiswahili|Somali|Oromo\\ 

19  & Global Voices & Text Summarization Dataset~\cite{nguyen2019global}  &  EMNLP & kiswahili\\ 

20  & Glot500 & Scaling Multilingual Dataset~\cite{imanigooghari2023glot500}& ACL & kiswahili|Somali|Luo\\ 

21  & LORELEI& disaster incidents Dataset~\cite{tracey2019corpus}   & DARPA & kiswahili|Somali|Oromo\\ 

22  & SIB-200& Topic classification Dataset~\cite{adelani2023sib}& ACL & kiswahili|Kikuyu|Dholuo\\ 

23  & Kamba POS Tagger Memory Based& POS Dataset~\cite{kituku2015kamba}  & IJNLC & kamba\\ 

24  & OSCAR & Common Crawl Dataset~\cite{abadji2022towards}  & ACL & Kiswahili|Somali\\ 

25  & Language modeling & Language modeling Dataset~\cite{shikali2019language}  & OSCAR & Kiswahili\\ 

26  & Spell-checker for Gikuyu & Spell-checker Dataset~\cite{chege2010developing}  & UON & Kikuyu\\ 

27  & CommonCrawl & cross-lingual Dataset~\cite{conneau2019unsupervised}   & ACL & Kiswahili|Somali|Oromo\\ 

28  & AI4D -African Language Program & cross-lingual Dataset~\cite{siminyu2021ai4d}   & IDRC & Kiswahili\\

29  & AfriMTE and AfriCOMET & human evaluation Dataset~\cite{wang2023afrimte}   & NAACL & Kiswahili\\

\bottomrule
\end{tabular}

\label{tab:text_speech dataset}
\end{table*}

This section provides an overview of publicly available datasets in the specified languages, focusing on their multilingualism characteristics, encompassing speech and text modalities. The datasets are further analyzed based on their applicability to various downstream tasks, highlighting their relevance for computational linguistics and machine learning applications.

\subsection{Speech Datasets}
\subsection*{(a) Kenyan Languages Corpus for Machine Learning and Natural Language Processing (Kencorpus):} KenCorpus\footnote{\url{https://kencorpus.maseno.ac.ke/corpus-datasets/}} \cite{wanjawa2022kencorpus} is a comprehensive text and speech corpus for three Kenyan languages: \textbf{Swahili}, \textbf{Dholuo} and \textbf{Luhya  (covering the Lumarachi, Logooli, and Lubukusu dialects).} The dataset was collected and curated using contributions from native speakers across diverse sources, including language communities, schools, media outlets, and publishers. KenCorpus provides a rich resource for machine learning (ML) and NLP applications; specifically, it comprises 2,585 texts  (approximately 1.8 million words) and 19 hours of speech for Swahili, 546 texts ( about 1.3 million words) and 99 hours of speech for Dholuo, and 987 texts (about 2.2 million words) and 58 hours of speech for the three Luhya dialects.  Additionally, the dataset includes KenSpeech, a transcribed corpus of roughly 27 hours of Swahili speech \cite{awino2022phonemic}. KenCorpus also features Part-of-Speech (POS) tagged sentences, covering approximately 50,000 words for Dholuo and 90,000 words for Luhya, further enhancing its utility for linguistic and computational research.

 
\subsection*{(b) Building African Voices:} The Building African Voices\footnote{\url{https://www:africanvoices:tech/ }} \cite{ogayo2022building} is a curated text and speech dataset with 16 African languages featuring 4 Kenyan languages: \textbf{Dholuo, Suba} and \textbf{Kenyan English}. The dataset contains 13,897 utterances in Luo, 2,078 in Suba, and 1,150 in Kenyan English. Data in this corpus was scraped from books, websites, and social media posts.

\subsection*{(c) CMU Wilderness Multilingual Speech Dataset:} 
CMU Wilderness\footnote{\url{http://festvox.org/cmu_wilderness/}} \cite{CMUWilderness_8683536} is a multilingual speech dataset containing several audio and textual data collected from the Bible. The dataset comprises audio in 700 different languages, with three (3) Kenyan languages represented, \textbf{Oromo, Somali} and \textbf{Sabaot}. On average, each dataset contains about 20 hours of aligned sentence-level text and word pronunciations. 

\subsubsection*{ (d) Bible TTS: }
The Bible TTS \footnote{\url{https://masakhane-io.github.io/bibleTTS/}}  is a speech dataset for 10 languages spoken in Sub-Saharan Africa, featuring \textbf{Kikuyu} and \textbf{Dholuo} languages \cite{meyer2022bibletts}. The dataset includes Bible recordings released by the Open.Bible project\footnote{\url{https://bibleproject.com/}}. The corpus contains up to 86 hours of high quality, aligned single speaker recordings per language.

\subsubsection*{(e) AfriSpeech-200} AfriSpeech \cite{olatunji2023afrispeech}, a Pan-African accented English speech dataset for clinical and general domain ASR, crowdsourced from 2,463 African speakers, 200.70 hrs with an average audio duration of 10.7 seconds. Kenya contributes 8,304 clips from 137 speakers, totaling 20.89 hours of audio. This data helps improve ASR performance for Kenyan-accented English, ensuring more accurate and inclusive speech recognition tools that cater to local healthcare systems and other domains where speech recognition can enhance productivity and accessibility.

\subsubsection*{(f) AfroDigits: }
This is a Community-Driven Spoken Digit Dataset\footnote{\url{https://huggingface.co/datasets/chrisjay/crowd-speech-africa}} \cite{chinenye2023afrodigits} for African. It is an openly available dataset of spoken digits in 38 African languages, including \textbf{Oromo} and \textbf{Borana}, spoken widely in the northern region of Kenya.

\subsubsection*{(g) 1000 African Voices: }
The Afro-TTS dataset\cite{ogun20241000}  is a collection of English speech recordings featuring diverse African accents, curated through crowdsourcing. It includes 136 hours of audio from 747 contributors across 9 countries, with 86 different accents. The Kenyan subset comprises 5,307 samples from 58 speakers, highlighting the unique linguistic characteristics of Kenyan English. This open-source dataset aims to support research on African-accented English.

\subsubsection*{(h) Building Text and Speech Datasets for Low Resourced Languages: A Case of Languages in East Africa: } This corpus \cite{babirye2022building} contains text and speech data in 5 languages including \textbf{Kiswahili}. Data was collected from contributions by the community and pre-existing text data sources: news websites, published reports, storybooks and open-source Kiswahili datasets. The Kiswahili dataset has over 200K sentences and 100 hours of voice data crowdsourced using Mozilla Commonvoice\footnote{\url{https://commonvoice.mozilla.org/en}}.

\subsubsection *{(i) XTREME-S: Evaluating Cross-lingual Speech Representations: }
The XTREME-S project\footnote{\url{https://hf.co/datasets/google/xtreme_s}} \cite{conneau2022xtreme}  created a speech dataset covering 102 languages, including  \textbf{Somali, Swahili, Dholuo} and \textbf{Kamba} from Kenya. \textbf{HOW was is created, QUANTITY} 

\subsubsection *{(j) The African Storybook Project  (ASb) : }
ASb\footnote{\url{https://www.africanstorybook.org/}} \cite{stranger2017african} is a children's literacy project by the South Africa Institute of Distance Education\footnote{\url{https://www.devex.com/organizations/south-african-institute-for-distance-education-saide-22060}}. The project is a collection of 4,317 storybooks, poems, songs, rhymes, and picture books in 242 African languages including \textbf{Ekegusii, Gikuyu, Dholuo, Turkana, Somali, Oromo, Maasai, Samburu} and \textbf{Kipsigis} languages spoken in Kenya. The textual data is organized according to the size of words and paragraphs.

\subsubsection * {(k) Mozilla Common Voice : }
Common voice\footnote{\url{https://commonvoice.mozilla.org/en}} \cite{ardila2019common} is the largest open-source, multi-language speech dataset. The text and speech datasets are collected and validated through crowdsourcing. It contains datasets in several languages, including 973 hrs of speech and 101,669 sentences in \textbf{Swahili}. This constantly scaling dataset also has 266 sentences in \textbf{Somali}. 



\subsubsection * {(k) IARPA Babel Dholuo Language Pack: }
 This is a speech dataset\footnote{\url{https://abacus.library.ubc.ca/dataset.xhtml?persistentId=hdl:11272.1/AB2/HSAU9N}}  \cite{AB2_HSAU9N_2022} containing approximately 204 hours of audio data in South Nyanza and Trans-Yala \textbf{Dholuo} dialects. The data is sourced from telephone speech. 

\subsubsection* {(m)A Kiswahili TTS dataset: }
This dataset\footnote{\url{https://data.mendeley.com/datasets/rbn6nmygcn/1}} \cite{Rono_2021}, available on Mendeley Data, contains 1,570 text files (23,487 words) and 1,570 audio files in \textbf{Kiswahili}. The textual data was sourced from newspaper articles, stories, and novels. 

\subsubsection* {(n) Swahili audio mini-kit: }
This dataset\footnote{\url{https://gamayun.translatorswb.org/download/swahili-audio-mini-kit/}} contains  4,700 samples from \textbf{Swahili} mini-kit recorded by a Kenyan male speaker and their transcriptions.

\subsubsection* {(o) Bloom-lm: } The Bloom library\footnote{\url{https://huggingface.co/datasets/sil-ai/bloom-lm}} \cite{leong2022bloom} has stories in 363 languages including the Kenyan languages \textbf{Taveta, Samburu, Swahili, Rendile, Ekegusii, Nyole, Turkana, Suba, Kikuyu, Oromo, Okiek, Kitharaka, Marakwet, Bukusu, Kamba, Pokomo} and \textbf{Meru}. The library, which has a mean of 32 stories and a median of 2 stories per language, was created by SIL International\footnote{\url{https://www.sil.org/}} to empower communities with low-resource languages to create literature for children.

\subsection{Text Datasets}

\subsubsection* {(a) Helsinki Corpus of Swahili (HCS): }
HCS \cite{hurskainen2004helsinki} is the most refined \textbf{Swahili} corpus containing 25 million words collected from news sources, stories, and legislative assemblies. The corpus is available in the Language Bank of Finland (Kielipankki) in two versions: annotated and not annotated. The not-annotated version is available openly via \url{https://korp.csc.fi/download/HCS/na-v2/ }. In contrast, the annotated version, which contains words annotated with individual \textbf{lemma, Part of Speech (PoS), morphological and syntactic tags} \cite{steimel2023towards} is access-restricted via \url{http://urn.fi/urn:nbn:fi:lb-201608301. }.

\subsubsection* {(b) MasakhaNEWS: } 
MasakhaNEWS \cite{adelani2023masakhanews} is the largest news classification dataset covering 16 African languages, including three languages in Kenya. The dataset has 7782, 6431, and 2915 news articles in \textbf{Oromo}, \textbf{Swahili}, and \textbf{Somali}, respectively. The news articles were analyzed in six categories: business, entertainment, health, politics, sports, and technology.

\subsubsection* {(c) AFRIHG: } 
AFRIHG \cite{ogunremiafrihg} is the largest abstractive summarization dataset for headline generation covering 16 African languages, including three languages in Kenya. The dataset has 11,998, 17,931, and 20,276 new articles and their respective headlines in \textbf{Somali}, \textbf{Oromo}, and \textbf{Swahili}, respectively. The dataset was scrapped from BBC News and incorporates news articles from \cite{adelani2023masakhanews}. 

\subsubsection* {(d) Global Voices } \cite{nguyen2019global} is a multilingual collection for evaluating cross-lingual summarization methods. It includes articles and their summaries in 15 languages. The dataset contains news articles and summaries in \textbf{Swahili}. The summaries are gathered automatically from social network descriptions and manually through crowdsourcing, ensuring quality through human ratings.

\subsubsection* {(e) Glot500: Scaling Multilingual Corpora and Language Models to 500 Language: }
The Glot500\footnote{\url{https://github.com/cisnlp/Glot500.}} project \cite{imanigooghari2023glot500} is an effort to scale NLP to support as many of the world's languages and cultures as possible. It created an LLM trained on 700GB of text in 2266 low-resource 'tail' languages, including \textbf{Swahili, Dholuo, Somali}  and \textbf{Oromo} from Kenya. The text data was collected by crawling websites and compiling data from around 150 different datasets.

\subsubsection* {(f) Corpus Building for Low Resource Languages in the DARPA LORELEI Program: } The LORELEI (Low Resource Languages for Emergent Incidents)\footnote{\url{https://www.darpa.mil/program/low-resource-languages-for-emergent-incidents}} program \cite{tracey2019corpus} developed large volumes of both monolingual and parallel language packs to improve technologies capable of providing awareness in disasters or emergent incidents. The two language packs, representative and incident, focus on low-resource languages and cover 23 languages, including \textbf{Oromo, Somali} and \textbf{Swahili}. The project's target was over 1 million words for the parallel datasets and over 2 million for the monolingual datasets. Data was sourced from formal news, social media, blogs, discussion forums, and reference materials such as Wikipedia. The textual data contains semantic, morphosyntactic, entity, and parallel annotations.

\subsubsection* {(g) SIB-200: A Simple, Inclusive, and Big Evaluation Dataset for Topic Classification in 200+ Languages and Dialects : }
SIB-200\footnote{\url{https://github.com/dadelani/sib-200}} \cite{adelani2023sib} is a benchmark topic classification dataset based on the FLORES-200 corpus that covers over 200 languages and dialects. The dataset contains  1,004 sentences and is annotated at sentence level in seven categories: science/technology, travel, politics, sports, health, entertainment, and geography. It contains data from Kenya in \textbf{Swahili, Gikuyu} and \textbf{Dholuo} languages.

\subsubsection* {(h) Kamba Part- of-Speech Tagger Using Memory Based Approach: }
This corpus contains approximately 30K words manually annotated with PoS tags. The words in the dataset were collected from online sources and documents written in  \textbf{Kikamba} language \cite{kituku2015kamba}.

\subsubsection* {(i) OSCAR: Open Super-large Crawled Aggregated corpus: }
The OSCAR project\footnote{\url{https://oscar-project.org/}} \cite{abadji2022towards} provides open-source resources and large, unannotated datasets. The current version contains web data in 166 languages, including 1,670 documents and 164,510 words in \textbf{Swahili} and \textbf{Somali}. 

\subsubsection* {(j) Language modeling data for Swahili: }

This dataset\footnote{\url{https://zenodo.org/records/3553423}} contains 28,000 unique words and a total of 9.81million \textbf{Swahili} words \cite{shikali2019language}. It was developed specifically for language modeling tasks.

\subsubsection*{(k) Developing an Open source Spell-checker for Gikuyu: } 
This corpus \cite{chege2010developing} is a collection of 19,000 words derived from pre-existing datasets in \textbf{Gikuyu}. It contains text from religious material, poems, short stories, novels, and the Internet. The data was manually annotated into parts of speech (PoS).

\subsubsection*{(l) Unsupervised Cross-lingual Representation Learning at Scale: } 

The study \cite{conneau2019unsupervised} resulted in CommonCrawl\footnote{\url{https://commoncrawl.org/}} Corpus\footnote{\url{https://github.com/facebookresearch/(fairseq-py,pytext,xlm)}}in 100 languages including \textbf{Oromo, Swahili} and \textbf{Somali}. This large corpus has over 2.5GB of data with 8 million tokens in Oromo, 275 million in Swahili, and 62 million in Somali.

\subsubsection*{(m) Building a database for Kiswahili language in Africa: } 
This document classification dataset contains 10K instances in \textbf{Swahili} and is a product of AI4D – African Language Program \footnote{\url{https://www.k4all.org/project/language-dataset-fellowship/}}\cite{siminyu2021ai4d}  which is working towards boosting the integration of African languages on digital platforms.

\subsubsection*{(n) AfriMTE and AfriCOMET: Enhancing COMET to Embrace Under-resourced African Languages: } 
\cite{wang2023afrimte} enhances machine translation (MT) evaluation for under-resourced African languages. It addresses the limitations of n-gram metrics like BLEU and introduces AfriCOMET, an evaluation metric with a higher correlation to human judgements. The dataset includes high-quality human evaluation data using simplified guidelines for error detection and direct assessment (DA) scoring across 13 diverse African languages, including Kenya; the language used is \textbf{Swahili}.

\subsection{Parallel Corpora}

\begin{table*}[ht]
\caption{Published Kenyan Parallel, Question and Answer, Sentiment Datasets, Hate Speech and NER Datasets} 
\footnotesize
\centering
\begin{tabular}{p{0.3cm} p{5cm} p{5cm} p{1.5cm} p{3cm}} 
\toprule
\# & \textbf{Dataset Name}     & \textbf{Available}    & \textbf{Venue}      & \textbf{Language}            \\ 
\midrule

1  & ParaCrawl: Web-Scale Acquisition & Parallel
Dataset~\cite{banon2020paracrawl} & ACL & Swahili|Somalia  \\

2  & MAFAND-MT & MT Parallel Dataset~\cite{adelani2022few} & ACL & Swahili|Luo\\ 

3 & WikiMatrix & Parallel Dataset~\cite{schwenk2019wikimatrix} & ACL & Swahili \\ 

4 & KenTrans &  Parallel Dataset~\cite{wanjawa2022kencorpus}& Maseno & Swahili|Luo|Luhya\\ 

5  & PanLex & lexemes pairwise Dataset~\cite{kamholz2014panlex}& LREC  & Kiswahili|Luhya|etc \\ 

6 &  FLORES-200 &  MT Parallel Dataset~\cite{conneau2022xtreme} & ACL & Swahili|Kikuyu|etc \\                   

7 & NLLB & multilingual African Stories~\cite{costa2022no} & Meta & Kikuyu|Swahili|etc \\ 

8  & CCAligned & Parallel Dataset~\cite{elkishky_ccaligned_2020}  & ACL  & Swahili|Oromo \\

9 & GoURMET   & Parallel speech~\cite{espla2019global} & \ ACL  & Swahili \\ 

10  & The SAWA & Parallel Dataset~\cite{de2009sawa}  & EACL & Swahili \\  

11  &  A Knowledge-Light Approach & Trillingual Dataset~\cite{de2010knowledge} & LREC & Luo \\ 

12  &  Tatoeba & Parallel Dataset~\cite{raine2018building}& Helsinki & Swahili|Somali|Rendile \\ 

13  &  Very LR Sentence Alignment & Parallel Dataset~\cite{chimoto2022very} & ACL& Swahili|Luhya \\

14  & English–Bukusu Automatic MT  & Parallel Dataset~\cite{ngoni2022english} & UON & Bukusu|Kenyan-eng \\ 

15  &  TICO-19 & Pairwise Dataset~\cite{anastasopoulos2020tico} & ACL & Swahili|Somali \\

16  & Tanzil  & Parallel Dataset\footnote{\url{https://tanzil.net/trans/}} &  Quran & Swahili|Somali\\ 

17  & AfriCLIRMatrix & I and R Dataset~\cite{ogundepo2022africlirmatrix}& EMNLP & Swahili|Somali\\ 

18  & CIRAL & I and R Dataset~\cite{adeyemi2024ciral}& ACM & Swahili\\

19  & KenSwQuAD & Q and A  Dataset~\cite{wanjawa2022kencorpus} & ACM & kiswahili\\ 

20  & AfriQA &  cross-lingual Q and A Dataset~\cite{ogundepo2023afriqa}& EMNLP & Swahili\\ 

21  & TYDI& Q and A  Dataset~\cite{clark2020tydi} & TACL & Swahili\\ 

22  & AfriSenti & Sentiment Dataset~\cite{muhammad2023afrisenti} & EMNLP & Swahili|Oromo\\ 

22  & RideKE & Sentiment Dataset~\cite{shikali2019language}  & ACL & Swahili|Sheng\\ 

23  & Hate Speech & Hate speech Dataset~\cite{ombui2019hate}  & IEEE & Swahili|Kenyan-eng\\ 

24  & XTREMESPEECH & Hate speechl Dataset~\cite{maronikolakis2022listening}   & ACL & Swahili|Kenyan-eng\\ 

25  & MasakhaNER & NER Dataset~\cite{adelani2021masakhaner}   & TACL & Swahili|Luo\\

26  & Mining Wikidata & NER Dataset~\cite{saleva2021mining}   & AfricaNLP & Swahili|Oromo\\

25  & Cross-lingual Tagging and Linking & NER and entity linking Dataset~\cite{pan2017cross}   & ACL & Swahili|Kikuyu|Somali\\

\bottomrule
\end{tabular}

\label{tab:text_parallel dataset}
\end{table*}
\subsubsection*{(a) ParaCrawl: Web-Scale Acquisition of Parallel Corpora: }

ParaCrawl \footnote{\url{https://paracrawl.eu/ }} \cite{banon2020paracrawl} is the largest parallel corpora curated by crawling the websites. It comprises 223 million unique sentence pairs in 23 languages, including 132,517 and 14,879 sentences in \textbf{Swahili} and \textbf{Somali}, respectively.

\subsubsection*{(b) Masakhane Anglo \& Franco Africa News Dataset for Machine Translation: }

MAFAND-MT \cite{adelani2022few} is a translated news corpus of 16 African languages, including \textbf{Swahili} and \textbf{Luo} sourced from news websites from local newspapers. The dataset contains 31K and 872K sentences in Swahili and Luo, respectively. 

\subsubsection*{(c) WikiMatrix: Mining 135M Parallel Sentences in 1620 Language Pairs from Wikipedia: }
Wikimatrix \footnote{\url{https://github.com/facebookresearch/LASER/tree/main/tasks/WikiMatrix }} \cite{schwenk2019wikimatrix} is a parallel corpus mined from Wikipedia containing 135M parallel sentences for 1620 different language pairs. The textual data is in 85 languages, including \textbf{Swahili}.



\subsubsection*{(d) KenTrans: }
On top of the individual monolingual datasets, the Kencorpus \cite{wanjawa2022kencorpus} project has parallel corpora between \textbf{Swahili} and \textbf{Dholuo} ( 1,500 sentences) and between Swahili and Luhya (11,900 sentences).

\subsubsection*{(e) PanLex: Building a Resource for Panlingual Lexical Translation: }
PanLex\footnote{\url{http://panlex.org }} \cite{kamholz2014panlex} is a documentation of 20million lexemes in 9,000 language varieties .
It covers several Kenyan Languages, including \textbf{Swahili, Somali, Samburu, Kipsigis, Makonde, Turkana, Oromo, Kuria, Suba} and the \textbf{Luhya} dialects Bukusu and Lulogooli. 

\subsubsection*{(f) FLORES-200 Evaluation Benchmark for Low-Resource and Multilingual Machine Translation: }
The FLORES 200\footnote{\url{https://github.com/facebookresearch/flores}} \cite{goyal2022flores} is a high-quality dataset consisting of 3,001 sentences mined from English Wikipedia and professionally translated in 204 languages. Some Kenyan languages covered by the dataset are \textbf{Swahili, Gikuyu} and \textbf{Dholuo}.
 
\subsubsection*{(g) No language left behind: Scaling human-centered machine translation: }
NLLB\footnote{\url{https://github.com/facebookresearch/fairseq/tree/nllb}} \cite{costa2022no} The dataset has texts from \textbf{Somali, Oromo, Kamba, Gikuyu, Dholuo} and \textbf{Swahili} Kenyan languages.


\subsubsection*{(h) CCAligned: A Massive Collection of Cross-lingual Web-Document Pairs: }
CCAligned corpus \cite{elkishky_ccaligned_2020} comprises parallel web document and sentence pairs in 137 languages, including the Kenyan languages \textbf{Swahili} and \textbf{Oromo} aligned with English. The data contains over 100 million aligned documents created by performing language identification on raw web documents and can be used for machine translation tasks.

\subsubsection*{(i) Global Under - Resourced MEdia Translation (GoURMET): } The GoURMET \footnote{\url{https://gourmet-project.eu/}} project curated a parallel dataset \cite{espla2019global} in 16 languages including 3.5 million sentences in \textbf{Swahili}. The data was sourced from web crawls. Language pairing is done to and from English, depending on the task.

\subsubsection*{(j) The SAWA Corpus: a Parallel Corpus English - Swahili} SAWA corpus \cite{de2009sawa} is a parallel corpus of English - Swahili curated to bootstrap a data-driven machine translation system for \textbf{English - Swahili}. The corpus comprises 542.1K and 442.9K words in English and Swahili, respectively. The textual data was collected from various sources, including the New Testament section of the Bible, Quran, kamusi (dictionary), reports, translators and movie subtitles.

\subsubsection*{(k) A Knowledge-Light Approach to Luo Machine Translation and Part-of-Speech Tagging: } This study \cite{de2010knowledge} resulted in a trilingual corpus \textbf{English - Swahili - Luo (Dholuo)} curated by utilizing the New Testament data of the SAWA corpus \cite{de2009sawa} to construct a trilingual parallel corpus (English - Luo - Swahili). The dataset contains 192K 156K and 170K token counts in English, Swahili and Luo respectively. Data is annotated with Part of Speech (PoS) tags.

\subsubsection*{(l) Tatoeba: }
Tatoeba\footnote{\url{https://opus.nlpl.eu/Tatoeba.php}} is a collection of over 11million sentences and translations in 446 languages including \textbf{Swahili, Somali} and \textbf{Rendile} \cite{raine2018building}. 

\subsubsection*{(m) Very Low Resource Sentence Alignment: Luhya and Swahili: }
This is the first digital parallel corpus in Luhya-English for the \textbf{Marama} dialect. The corpus comprises 7,952 parallel sentences in Marama generated by aligning the Bible's New Testament in Luhya and English \cite{chimoto2022very}.

 \subsubsection*{(n) English–Bukusu Automatic Machine Translation for Digital Services Inclusion in E-governance: } This study resulted in a parallel \textbf {English-Bukusu} dataset containing 5,146 sentences. The data was collected from three sources: the New Testament section of the Bukusu version of the Bible, electronic texts in Bukusu and the English-Bukusu dictionary \cite{ngoni2022english}.

\subsubsection*{(o) Translation Initiative for COVID-19: } TICO-19 Translation Benchmark \footnote{\url{https://tico-19.github.io/}} is a translation of COVID-19-related terms from English to various languages. This benchmark aims to include 30 documents (3071 sentences, 69.7k words) translated from English into 36 languages, including \textbf{Somali} and \textbf{Swahili}. \cite{anastasopoulos2020tico}

\subsubsection*{(p) Tanzil Dataset : } Tanzil\footnote{\url{https://tanzil.net/trans/}} is a collection of Quran translations in 42 languages including \textbf{Swahili} and \textbf{Somali}.

\subsection{Question Answering Datasets}
\subsubsection*{(a) AfriCLIRMatrix: }
AfriCLIRMatrix\footnote{\url{https://github.com/castorini/africlirmatrix}} \cite{ogundepo2022africlirmatrix} is an information retrieval (Question Answering) test collection comprising queries and documents in 15 African languages including \textbf{Swahili} and \textbf{Somali}. This dataset, sourced from Wikipedia, contains 9860 documents from Somalia and 70808 from Swahili.

\subsubsection*{(b) Cross-lingual information retrieval: }
CLIR\footnote{\url{https://github.com/ciralproject/ciral}} \cite{adeyemi2024ciral} is an information retrieval (Question Answering) test collection comprising queries and documents in 15 African languages including \textbf{Swahili} and \textbf{Somali}. This dataset, sourced from Wikipedia, contains 9860 documents from Somalia and 70808 from Swahili.

\subsubsection*{(c) KenSwQuAD—A Question Answering Dataset for Swahili Low-resource Language: } 
KenSwQuAD \cite{wanjawa2023kenswquad} is a Question Answering dataset containing about 1,440 \textbf{Swahili} texts labeled with at least 5 questions each, totaling about 7,526 QA pairs. The dataset is based on the KenCorpus \cite{wanjawa2022kencorpus} project. 

\subsubsection*{(d) AfriQA: Cross-lingual Open-Retrieval Question Answering for African Languages: }
AfriQA\footnote{\url{https://github.com/masakhane-io/afriqa}}  \cite{ogundepo2023afriqa} is the first cross-lingual QA dataset containing 10 African languages, including 1,134 questions in \textbf{Swahili}. Native speakers of each language were tasked with data collection and annotation. The process involved question elicitation, translation into a pivot language (English or French), answer labelling and answer translation back to the source language.

\subsubsection*{(e) TYDI QA: A Benchmark for Information-Seeking Question Answering in Typologically Diverse Languages:} TYDI QA\footnote{\url{https://github.com/google-research-datasets/tydiqa}}  \cite{clark2020tydi} is a QA dataset covering 11 languages including \textbf{Kiswahili}. The dataset contains 204K QA pairs collected by generating questions based on short prompts from Wikipedia articles and then pairing each question with a Wikipedia article.

\subsection{Sentiment Analysis}
\subsubsection*{(a) AfriSenti: }
Afrisenti \cite{muhammad2023afrisenti}  is a Twitter sentiment analysis dataset consisting of annotated textual data in 14 African languages, including \textbf{Swahili} and \textbf{Oromo} spoken in Kenya. The dataset contains 3,014 and 2,491 texts labelled as negative or positive in Swahili and Oromo, respectively.

\subsubsection*{(b) RideKE: } RideKE \cite{etori2024rideke}  is a Twitter-based corpus containing over 29,000 code-switched entries in Kenyan-accented English, Swahili, and Sheng. It is designed explicitly for sentiment and emotion analysis within the ride-hailing service domain. The dataset includes 553 labelled entries for supervised training, 2,000 human-annotated entries for testing, and over 27,000 unlabeled entries used in a semi-supervised learning loop.

\subsection{Hate Speech}
\subsubsection*{(a) Hate Speech Detection in Code-switched Text Messages: }
The study by \cite{ombui2019hate} curated a dataset with 260k tweets in \textbf{Swahili, English} and other Native African languages. The data was sourced from Twitter and is annotated in three categories: offensive, hate, and neither. 

\subsubsection*{(b) XTREMESPEECH: Listening to affected communities to define extreme speech: }
XTREMESPEECH\footnote{\url{https://github.com/antmarakis/xtremespeech}}  \cite{maronikolakis2022listening} is a hate speech dataset in 6 languages including \textbf{Swahili}. It contains 20,297 passages collected from social media and annotated in three categories: derogatory, exclusionary, and dangerous.

\subsection{Named Entity Recognition}
\subsubsection*{(a) MasakhaNER: }
MasakhaNER\footnote{\url{https://git.io/masakhane-ner}} \cite{adelani2021masakhaner}  is the first, publicly available Named Entity Recognition dataset in 10 African Languages sourced from local news. The dataset contains 921 and 3,006 sentences in \textbf{Luo} and \textbf{Swahili} respectively. The sentences are annotated in four categories: personal name, location, organization and date \& time.

 \subsubsection*{(b) Mining Wikidata for Name Resources for African Languages: }
This project\footnote{\url{https://github.com/bltlab/africanlp2021-wikidata-names}}~\cite{saleva2021mining} contains a list of approximately 1.9million names in 28 African languages mined from Wikipedia. It contains 23,0614 \textbf{Swahili} names, 26,215 \textbf{Oromo} names and 28,215 \textbf{Somali} names. 

\subsubsection*{(c) Cross-lingual Name Tagging and Linking for 282 Languages: } The WikiAnn project\footnote{\url{https://elisa-ie.github.io/wikiann/}} developed cross-lingual name tagging and linking for 282 languages in Wikipedia including the Kenyan languages \textbf{Swahili, Kikuyu, Somali} and \textbf{Oromo}. The data is annotated for three entity types: PER, ORG and GPE/LOC. It contains 9.3K, 6.5K, 1.0K and 709 names Swahili, Somali, Kikuyu and Oromo respectively \cite{pan2017cross}.

\subsection{Dictionaries}
\subsubsection*{(a) Glosbe Dictionary: }
Glosbe\footnote{\url{https://glosbe.com/}} is the most extensive online community-built dictionary that provides free dictionaries with in-context translations. The dictionary supports 6,000 languages, including the Kenyan languages \textbf{Swahili, Kikuyu, Luo, Gusii, Kalenjin, Meru} among others. Aside from sentence translation, the dictionary contains phrase illustrations, audio recordings and pronunciations, translated sentences, and automatic translators for long sentences. It has over 2 billion translations, 400K audio recordings, and 1 billion sentence examples.

\subsubsection*{(b) Mandla African language dictionary: }
Mandla\footnote{\url{https://dictionary.sebmita.com/}} is a free, multilingual dictionary that translates in over 100 African languages. The crowd-sourced dictionary, which has both text and audio options, gives both the native and Latin script definitions of words and has over 75K users. Some Kenyan languages Mandla supports are \textbf{Swahili, Kipsigis, Oromo, Makonde, Turkana, Kamba, Somali, Oromo} and \textbf{Dholuo}.

\subsubsection*{(c) A Lexicon of Key Words in Kiswahili: }
This is a dictionary\footnote{\url{https://pacscenter.stanford.edu/publication/a-lexicon-of-key-words-in-kiswahili/}} with 52 \textbf{Swahili} words and phrases relating to technology translated to English \cite{Nyabola_2022}. The dictionary was created to spark conversation on digital rights in languages other than English and is labelled according to Part of Speech.

\subsubsection*{(d) IPA-dict: Monolingual wordlists with pronunciation information in IPA: }
IPA-dict\footnote{\url{https://github.com/open-dict-data/ipa-dict}} is the first standardized series of dictionaries of wordlists with accompanying phonemic pronunciation information in IPA - International Phonetic Alphabet transcription \cite{doherty2019ipa}. The data is in 23 languages, including \textbf{Swahili}. 

\subsubsection*{(e) Common Swahili Slangs: }
This dataset\footnote{\url{https://data.mendeley.com/datasets/b8tc96xf3h/}} contains 188 Swahili slangs and their respective proper words \cite{Masasi_2020}.

%% file: applications.tex
\section{Applications} 
\label{secapplication}

This section provides an overview of various NLP tasks and the publicly available models for each. We also examine the current state of Kenyan languages about these tasks and techniques, highlighting the challenges and progress in developing NLP resources for these languages.

\subsection{Machine Translation}
\paragraph{Rule-Based approach:}
This is one of the earliest methods in the field based on a deep understanding of the linguistic properties of both source and target languages. It combines expert-crafted grammar rules and dictionaries, focusing on specific linguistic aspects such as morphology, syntax, and lexical semantics. All Kenyan languages are still considered low-resource; thus, they do not have curated grammar rules for rule-based machine translation. on the other hand, all languages listed in Table \ref{tab:main_languages} have available dictionaries online\footnote{can be found on this site: \url{https://glosbe.com/en}}. Collectively, even with these resources, are few to no publicly available rule based MT models for Kenyan languages.

\paragraph{Statistical Machine Translation:}
This approach employs statistical techniques such as probability models to facilitate translation between source and target languages. This approach is based on the analysis of large corpora of bilingual text data, where the system learns how words, phrases and sentences in one language correspond to those in another language. It assigns probability scores to words or phrases in each target sentence, with those scoring highest considered the best translations. De Pauw et al. \footnote{https://aclanthology.org/www.mt-archive.info/MTMRL-2011-DePauw.pdf}, developed a Swahili-English Statistical Machine translation system

\paragraph{Neural Machine Translation:} 
This approach is currently regarded as the state of art in the field as it has shown improved performance compared to the other techniques. It employs deep learning techniques to infer high-level semantics from language translations. A prominent method in this approach is the transformer-based model with encoder-decoder architecture introduced by Vashwan et al.(2017)
Many African languages including Kenyan languages have not benefited from these developments as parallel data is scarce for these languages. However, there have been several efforts to build NMT models or include them in multilingual MT models. We will break this down according to the various languages:
\begin{enumerate}
    \item Swahili: This  stands as the most resourced language in Kenya, prominently featured in both national and international contexts. Several models cover Swahili, namely: bilingual model, multilingual model, fine-tuned models. 
    
    \paragraph{Bilingual Models}:
    \begin{itemize}
        \item Rogendo's English-Swahili Model: Available at Hugging Face, this model represents a key resource for English-Swahili translation\footnote{https://huggingface.co/Rogendo/en-sw}.
        \item HPLT's Swahili-English Model: Trained using data from OPUS and HPLT, this model facilitates Swahili to English translation\footnote{\url{https://huggingface.co/HPLT/translate-sw-en-v1.0-hplt_opus}}.
        \item Helsinki's English-Swahili and Finnish-Swahili Models: These models are part of Helsinki-NLP’s offerings, demonstrating a broader linguistic reach\footnote{https://huggingface.co/Helsinki-NLP/opus-mt-en-sw}\footnote{https://huggingface.co/Helsinki-NLP/opus-mt-fi-sw}.
        \item Masakhane English-Swahili models: These models are found on GitHub and form part of the machine translation for Africa project.
    \end{itemize}
    \paragraph{Multilingual Models}:
    \begin{itemize}
        \item \textbf{NLLB, MMTAfrica, m2m-100:} These models incorporate Swahili as part of their multilingual capabilities, highlighting the language's significant presence in global language models.
    \end{itemize}
    \item Kikuyu: Integrated into the NLLB model, and subsequent fine-tuning efforts. Additionally, a Swahili-Kikuyu translation model is hosted on GitHub\footnote{\url{https://github.com/starnleymbote/Kikuyu_Kiswahili-translation}}, along with a Masakhane’s English-Kikuyu model.
    \item Kamba: Similarly to Kikuyu, included in both NLLB and Masakhane's models available on GitHub.
    \item Luo: Featured in multilingual models like NLLB and specific bilingual models by Helsinki-NLP\footnote{https://huggingface.co/Helsinki-NLP/opus-mt-en-luo}.
    \item Luhya:  Despite available parallel datasets, there are no publicly available machine translation models for Luhya.
    \item Somali, Oromo, Maasai, Nandi (Kalenjin): Included in the MADLAD-400 model, with Oromo supported by Helsinki's bilingual models and Somali featured in NLLB.
    \item Gusii, Meru, Giriama/Mijikenda, Turkana, Borana, Orma: No models are publicly available for these languages.
\end{enumerate}

\paragraph{Closed MT Systems}

There are several closed machine translation systems that feature Kenya languages. Namely: Namely: Google Translate features Swahili and Oromo, machinetranslate.org features Swahili, Kikuyu, Kamba, Somali and Oromo, rephrasely features Kikuyu.

\subsection{Information Retrieval}



Information retrieval is still an underexplored task in the Kenyan context. We see the growing research initiatives such as Masakhane\footnote{https://www.masakhane.io/}, and the advancement of deep learning methodologies such as transfer learning help in building new African datasets by adapting existing multilingual pretrained models in high-resource languages \cite{ogueji2021small,ogundepo2023afriqa} that promote improving natural language processing and information retrieval for African languages \cite{wanjawa2023kenswquad,abedissa2023amqa}. However, these previously mentioned works only cover Swahili, leaving opportunity for exploration for the rest of the Kenya languages.

\subsection{Text Classification}

\subsubsection{Sentiment analysis}

In this subsection, we review of the state of sentiment analysis in Kenya.

Sentiment analysis for Kenyan languages has been explored limitedly in multilingual settings. For instance, \cite{muhammad2023afrisenti} collected Afrisenti geographically distributed over the African continent  and included Oromo and Kiswahili, spoken in Kenya. Other sentiment analysis applications in Africa documented include Sentimentr \cite{cannon2022understanding} , VADER\cite{botchway2020deductions} and NVIVO II \cite{ochieng2017exploration}.

Although limited below are some sentiment analysis applications developed for the Kenyan context.

\begin{itemize}
    \item The mapping of sentiment of the Government of Kenya’s state service, \textit{Huduma Kenya} \cite{ng2018prototype}. Ng’angira develops a prototype system to obtain, preprocess and make inferences about public sentiment on government services. For a use case, Ng’angira tests the system on 8001 tweets; 3307, 2658 and 2036 positive and neutral respectively. 
    \item Sauer et al have analysed the Kenyan public’s reaction to China’s Belt and Road Initiative (BRI) \cite{morrison2022analyzing}. This is a project by China for the building of large scale infrastructure in 150 countries in Africa, Middle and Far East and Europe. In Kenya, the flagship project of the BRI is the Standard Gauge Railway (SGR).   The Authors use VADER to analyse a set of multilingual tweets in English and Kiswahili constituting a commentary corpus on the SGR. For Kiswahili, the authors first translate the tweets into English then carry the sentiment analysis on the English translation. For error analysis, 200 tweets were randomly sampled for each language and sentiment class. While there’s a worry that there could be an error amplification from translation, the author’s demonstrate that the Swahili and English error rates do not demonstrate significant divergences.
    \item  Moge Noor analyses sentiment reaction to 1122 tweets on the Demonetisation of legal tender using multinomial naive Bayes classification\cite{noor2020sentiment}.
    \item Evanega et al on the other hand compare the attitudes in Kenya and other African countries (Uganda, Nigeria)  towards the introduction of Genetically Modified Organisms (GMO) in the country \cite{evanega2022state}. From a baseline of around 67\% positive or neutral views towards GMO, there has been a slight increase towards positive attitudes towards GMO. They show that this level is relatively stable from January 2018 to December 2020. 
    \item \cite{noorSentimentAnalysis} developed a model to analyse sentiment of demonetization in Kenya. The data was retrieved from twitter using a web scraper with the help of advanced search. The data was collected between June and October , 1128 tweets collected. Some techniques for data processing were used to reduce noise and dimensionality in the data. The SA was performed using the Multinomial Naive Bayes Algorithm and the development of lexicon  analysis  which contains a word list that is negative and positive. For model evaluation, the author used a confusion matrix to summarize the performance of the prediction and its results.To validate the results, precision,recall and accuracy was used. The overall score maintained was an accuracy of 0.704.
    \item \cite{Jytte2015Tweeting} used sentiment analysis to help with identification, detection and tracking of terrorists activities in a more timely manner. Data was collected from twitter where the analysis was done. To accomplish the analysis ,analyst must have real time access to incident reports and make information scanning a daily process.
    \item \cite{Ngoge2015Mapping} carried out mapping of sentiments, which is the process of linking classified tweets and geocoordinates.

\end{itemize}

\subsection{Automatic Speech Recognition(ASR) }
There is a significant opportunity to create and expand datasets and applications in local languages and dialects to improve the accuracy of ASR. Increased collaboration between local universities, research institutions, and international organizations can accelerate advancements in ASR technology. ASR technology can be applied in public service sectors such as education, healthcare, and government to improve accessibility and efficiency. Encouraging local startups to innovate in the ASR space can lead to the development of tailored solutions that address specific regional needs \cite{asr_opportunities1}. However, as it stand, there are limited ASR applications for Kenyan languages.

\subsection{Named Entity Recognition} 

We detail NER contributions in Kenya in this subsection:

The Swahili NER dataset is a Named Entity Recognition (NER) dataset generated from https://huggingface.co/datasets/swahili using back-translation techniques.
This data has been cleaned using a couple of techniques and is ready for training a Spacy NER model without any modifications, with this data one is able to train a swahili-spacy-ner contributing to the NER space in Kenya.

Another application of NER is the Analysis of a machine learning-based algorithm used in name entity recognition. The work analysed various machine learning algorithms and implemented KNN which has been widely used in machine learning and remains one of the most popular methods to classify data.  It was established by the researchers that no published study has presented Named entity recognition for the Kikuyu language using a machine learning algorithm. This research will fill this gap by recognising entities in the Kikuyu language. 
An evaluation was done by testing precision, recall, and F-measure. The experiment results demonstrate that using K-NN is effective in classification performance compared to other models such as Naive Bayes,Maximum Entropy Model,Support Vector Machine (SVM),Decision Tree Learning and Hybrid.With enough training data, researchers could perform an experiment and check the learning curve with accuracy that compares to state of art NER.\cite{Kamau2023}






\subsection{Question Answering} 
The application of question-answering methodologies to indigenous Kenyan languages significantly lags behind rich-resource languages such as English. Recently, advancement have relied on large-scale language modeling, for which one example is available in Kiswahili, UlizaLlama, by JacarandaHealth, which uses a version of the ALPACA dataset translated into Kiswahili. version of the ALPACA dataset. Another initiative aimed at advancing question-answering for African languages is AfriQA

The need for Question Answering datasets in low resource languages also motivated the development of KenSwQuAD.The dataset was annotated from raw story texts of Swahili low resource language, which is a predominantly spoken in Eastern African and in other parts of the world. Question Answering (QA) datasets are important for machine comprehension of natural language for tasks such as internet search and dialogue systems. The research involved annotators in creating QA pairs from Swahili texts gathered by the Kencorpus project, which is a corpus of Kenyan languages. Out of a total of 2,585 texts, 1,445 were annotated, each with at least five QA pairs, resulting in a final dataset of 7,526 QA pairs. A quality assurance check on 12.5 percent of the annotated texts verified the correctness of all QA pairs. A proof of concept demonstrated that this dataset is suitable for QA tasks. Additionally, KenSwQuAD has enhanced the resources available for the Swahili language.
The dataset used for generating
KenSwQuAD was the Swahili portion of the data collected by the Kencorpus project\cite{Wanjawa2022}. Kencorpus project collected
primary data, both text and voice, in three low-resource languages of Swahili, Dholuo and Luhya.



\subsection{Natural Language Inference (NLI) or Textual Entailment: }
Natural Language Inference (NLI), or textual entailment, is a fundamental NLP task determining whether a given hypothesis logically follows a provided premise. This task typically classifies the relationship between the premise and hypothesis into one of three categories: entailment, contradiction, or neutral. While NLI models are well-developed for high-resource languages like English, their application to low-resource languages, such as those spoken in Kenya, remains underexplored. Most NLI models that support Kenyan languages, particularly Swahili, are multilingual. These models often rely on translations of English NLI datasets into Swahili. One example is the Multilingual NLI dataset \footnote{\url{https://huggingface.co/datasets/MoritzLaurer/multilingual-NLI-26lang-2mil7}}, which supports Swahili and has been used in models such as \texttt{mDeBERTa-v3-base-xnli-multilingual-nli-2mil7}. This model provides coverage for Swahili, leveraging cross-lingual capabilities to address the lack of Swahili-specific NLI datasets. Additionally, the afriXNLI dataset includes a test set for various African languages, though Swahili is the only Kenyan language represented in this dataset. Other Kenyan languages, such as Kikuyu, Luo, and Luhya, currently lack dedicated NLI datasets or models, which presents a significant gap in the research landscape.
\subsection{Pretrained Language Models (PTMs)}

In the AI revolution, language modeling is fundamental to NLP. At its core, language modeling involves prediction of the next word in a sequence given a probability distribution. These models are used in various applications from autocomplete to more complex tasks like classification and machine translation. The growth of language models has been catalysed by the availability of large datasets of textual data and development of deep learning architectures such as transformers.

Despite these advancements in language modelling, African languages have lagged behind due to various factors. Namely: data scarcity, technological infrastucture and commercial interest. Below we discuss the coverage of Kenyan languages in pretrain language models and how they are being utilised.

\begin{itemize}
    \item BERT-models, GPT-models, BLOOM-models, Claude, LLaMA-models.
    
BERT -models which was developed by Google, has been widely used in language modelling . It currently supports 100 languages however only one Kenyan language, Swahili, is supported.\cite{google2024bert}
The GPT models by OpenAI doesn't support Kenya languages yet however it understands around 60 indigenous languages from Kenya with limited proficiency.
BLOOM-models developed by The BigScience project supports 46 natural languages and of those Swahili is the only Kenyan language supported with a percentage contribution of 0.02.\cite{bloom_training_data}
Claude developed by Anthropic is known to include support for widely spoken African languages such as Swahili, which is a major language in Kenya however there is no mention of other Kenyan languages.
LLaMA-models , support several Kenyan languages, focusing primarily on Swahili. Swahili is one of the most widely spoken languages in Kenya. LLaMA models aim to extend support to other underutilized Kenyan dialects, although the specifics about these dialects are not detailed.
    \item Africa AI o-specific LMs (AfriBERTa, Afro-XLMR, AfroLM, SERENGETI, mBERT)

AfriBERTa \cite{ogueji2021small} model was trained on 11 languages which includes support for Swahili and Somali, two major languages spoken in Kenya.

AfroXLMR \cite{alabi2022adapting} model was trained on 17 languages which includes representation of Swahili and Somali which is spoken in Kenya.

AfroLM \cite{dossou2022afrolm} model is trained on 23 African languages including the Kenyan languages Luo and Swahili.

The SERENGETI \cite{adebara2022towards,adebara2022serengeti} is a multilingual model covering 517 African languages including Luo, Bukusu, Embu, Chidigo, Ekegusii, Oromo and Borana-Arsi-Gurji, Gikuyu, Kalenjin, Kikamba, Kuria, Maasai, Giryama, Pokomo (Kipfokomo), Rendile, Samburu, Somali, Swahili, Teso, Turkana and Tharaka. Datasets in the model are classified as Nilo-Saharan, Niger-Congo or Afro-Asiatic

    \item LMs finetuned on Kenyan specific languages (UlizaLLaMA, SwahBERT). These models have been fine-tuned with Swahili data. 

\textit{\textbf{Application Areas of PTMs} }

UlizaLLaMA, developed by Jacaranda Health \footnote{\url{https://jacarandahealth.org/jacaranda-launches-first-in-kind-swahili-large-language-model/}}, is finetuned to converse fluently in Swahili. This model helps to improve healthcare. 

The AfroXLMR model has been used by \cite{muhammad2023semeval} in sentiment analysis for 14 African languages including Swahili and Oromo. The model is able to perform both monolingual and multilingual classification of text in three classes: positive, negative or neutral. 

The study by \cite{alabi2022adapting} used AfriBERTa in multilingual adaptive finetuning (MAFT) for 17 African languages including Oromo, Swahili and Somali. In the Kenyan context, this model can be used in cross-lingual transfer learning.

mBERT which is a BERT model has multilingual capabilities to handle tasks such as Name Entity Recognition for African languages as seen in \cite{adelani2021masakhaner}. The model covers 10 African languages including Swahili and Luo.

\end{itemize}


%% file: governance.tex
\section{Governance, Policies, and Regulations} 
\label{secgov}

Governance, policy, ethics, and responsible NLP are interconnected concepts that relate to the oversight, regulation, and moral considerations of applying Natural Language Processing technologies. \citet{siau2018} studied the general concepts of AI governance, policies, and regulations. In their work, they argue that AI and other related technologies are expeditiously advancing. Hence, there is a need to discuss governance, policy, and regulatory issues in the field. The paper calls on relevant researchers, policymakers and government officials to take these matters into consideration. 

\subsection{Assessing AI governance}

Progress on  AI governance, NLP included, has been slow not only in Kenya but also in the wider continent. The ALT Advisory 2022  report \cite{alt_advisory} on AI governance, provides a useful set of indicators upon which to assess continental and country-level progress on AI governance: existence of AI legislation, existence of Data protection integrating rights on automated decision making, existence of a national AI strategy,existence of a draft policy on AI and presence of a task force / commission to oversee AI adoption and AI integration in the national development plan.

\subsection{State of AI governance in Kenya}
\subsubsection{Legislation and strategy}

Kenya, like many countries in Africa, neither has a strategy nor regulation governing the AI sector \cite{ai_strategy}. That said,the past year saw significant legislative activity.On September 8, 2023 The Ministry of Information, Communications, and the Digital Economy unveiled  the Working Group on Policy and Legislative Reforms for the Information, Communications and the Digital Economy Sector. The mandate of the sector working group includes reviewing current policy,legal and institutional structures, identifying emerging  technologies requiring oversight  and making legislative and policy recommendations. As AI is encapsulated under emerging technologies, recommendations from this task force are of keen interest to stakeholders. Regarding AI regulation, the Robotics Society Of Kenya (RSK) introduced a draft bill seeking to regulate the Robotics and AI sector. The draft titled the "Kenya Robotics and Artificial Intelligence Society Bill, 2023" seeks to provide for the establishment of the Kenya Robotics and Artificial Intelligence Society whose envisioned mandate includes regulation,licensing,policy guidance,awareness creation and capacity building \cite{robotics}.Overall, the draft bill, currently in the petition stage, has attracted vehement opposition from practitioners on the basis of being punitive, bureaucratic and ill-timed.While it remains to be seen whether the draft bill will  proceed beyond the petition, its mark on the governance discourse , albeit controversial, is cemented.

\subsubsection{Data protection integrating rights on automated decision making}

In 2019,  the Data Protection Act No. 24 of 2019 came into effect \href{https://www.kentrade.go.ke/wp-content/uploads/2022/09/Data-Protection-Act-1.pdf}{}. Although not explicit on AI governance, it is of great relevance to stakeholders within the AI ecosystem, given that data is the cornerstone of AI technology development. Key tenets of the act related to data protection include: 
\begin{itemize}
    \item incorporating operational and technical systems that include data protection principles, enforceability mechanisms, risk management, cyber-security measures, access security, physical security, and de-identification measures in all AI software and applications like chatbots,
    \item implementing security safeguards to ensure that personal data is accessed only by authorised persons. This includes technical safeguards for encryption, personnel vetting, and procedural safeguards like restricted access control and continuous database backup,
    \item limiting the amount of personal data collected for a given purpose. Thus, avoiding the processing of personal data altogether when possible and ensuring the data collected is minimised, 
    \item enforcing transparency and lawfulness. That is, use clear and plain language to communicate with data and related subjects, 
\end{itemize}

\subsubsection{Presence of task force, commission to oversee AI adoption}

Increasing interest in the opportunities and challenges brought about by AI technologies has elicited significant interest from stakeholders within the AI ecosystem, leading to the formation of task forces with sector-specific mandates. To illustrate, the Kenya Blockchain and AI Task Force, formed in 2018, was mandated to study emerging technologies, their use cases for development goals, and opportunities for legislation \cite{emerging_technologies}. The task force outlined important use cases for AI technologies within Kenya's development framework, and these suggestions are officially integrated into Kenya's economic development blueprints through the following:

\begin{itemize}
    \item \textbf{Data Science Africa (DSA)} - This is a non-profit organisation founded in 2019 by Kenyan researchers, practitioners, and entrepreneurs passionate about using AI to solve real-world problems. It is main mission is to work with institutions of higher learning and other government authorities to promote AI and data science for academia and R\&D, and support the adoption of data science and AI in businesses and government. The researchers are also advocating for ethical use of AI. DSA works with the government to influence AI policies. \cite{DSA}.
    
    \item \textbf{Information \& Communications Technology Authority (ICTA)} - I was established in 2013. It is advocating for e-Government services, development and use of ICT, efficient and effective provision of ICT services and applications, rights of ICT users,and cyber-security. The authority is working to streamline the management of AI and ICT functions within the national and county governments \cite{icta}. 
    
    \item \textbf{Office of the Data Protection Commissioner (ODPC)} - This is government body that is entrusted safeguard personal data. It was established under the Data Protection Act of parliament in 2019. Its mandates are available in \cite{odpc} and on this bases advocates for safe and ethical use of private data in AI research and innovation. 
    
    \item \textbf{The Communications Authority of Kenya (CA)} serves as the regulatory body that oversees the country's communications sector. It was founded in 1999 under the Kenya Information and Communications Act of 1998. Its mandate covers the advancement of various segments within the fields of information and communication, such as broadcasting, cyber security, multimedia, telecommunications, e-commerce and post and courier services \cite{cak}. In mid-2024, that authority operationalized the regulatory sandbox to test, monitor, and govern emerging technologies, innovations, and applications for AI-driven services, including chatboxes and BERTs \cite{sandbox}.
\end{itemize}
Both ICTA and CA are domiciled in Ministry of Information Communication and Technology.

%% file: indigenous.tex
\section{Indigenous Knowledge Systems and NLP}
Indigenous or local knowledge systems (IKS) refer to a body of knowledge, skills, innovations, value and belief systems passed down through generations in a given cultural locality and acquired through the accumulation of experiences, relationships with the surrounding environment, and traditional community rituals, practices and institutions \cite{marshall2019ai}. The indigenous knowledge systems are not mere collections of facts but are dynamic, living libraries of understanding, embodying the wisdom of centuries. IKS encompass a spectrum of skills, innovations, and deeply held beliefs rooted in the rhythm of local ecosystems, nurtured by communal rituals and enduring practices.
Natural Language Processing (NLP) stands as a bridge between this ancient wisdom and the digital age. It is a tool that has the potential to decode and digitise these verbal and non-verbal languages, transforming them into accessible data for the global stage. 

%% file: discussion.tex
\section{Discussion}
\label{secDiscussion}


Natural Language Processing (NLP) has garnered growing attention over the past decade, highlighting the undeniable need for continued research to enhance these techniques. This section provides insight and discusses the progress made in NLP in Kenya so far. 

Section \ref{secDataset} presents that the NLP ecosystem in Kenya is rapidly expanding. These can be understood in the context of the African continent's wider digitisation and technologisation. Progress in North, South, West, and East Africa has provided a context and complement to NLP efforts in Kenya. Internally, the development of datasets such as Kencorpus \cite{wanjawa2022kencorpus}, KenSwQuAD\cite{wanjawa2023kenswquad} and Mozilla Common Voice\footnote{https://commonvoice.mozilla.org/en} have helped create the necessary input infrastructure for applications such as the case in machine translation, information retrieval, sentiment analysis and in general language modelling tasks. As a consequence, various application models such as Serengeti \cite{adebara2022serengeti}, Afribert \cite{ogueji2022afriberta}, and AfroXLMR \cite{azime2023masakhane} have been developed for the Kenya NLP application contexts. 

This has been accentuated significantly by international work seeking to broaden the scope of open-source or commercial products to the African continent. International efforts in the creation of datasets in Swahili\cite{chimoto2022very}, Oromo\cite{pan2017cross}, Somali\cite{ogunremiafrihg} and other languages have triggered the digital enrichment of various Kenyan languages. Several languages in Kenya, including most recently Dholuo, have been incorporated into Google Translate \cite{goyal2022flores}.

However, this relatively rapid expansion of the NLP ecosystem in Kenya is not without qualification. From an inspection of the covered datasets, it can be seen that the language with the most resources is Swahili, followed by Oromo, Somali, and Luo. Most of these languages are transboundary, hence their attractiveness for digitisation. This is also reproduced in the wider continent, where Bambara (spoken in Mali, Ivory Coast and other West African Countries) \cite{diallo2021bambara}, Arabic in the North \cite{fourati2020tunizi}, and Nguni languages (IsiZulu) in the South are widely captured in NLP datasets\cite{marivate2020african}. There is a danger of "tail of tail" differentiation where the least resourced African languages in need of the most urgent digital preservation remain unpreserved, leading to a potential loss of culture in the continent.'

Concomitant to this, there is a possibility that these under-resourced languages will be considered as an international "terra nullius" - open for the preservation (and possible exploitation) by any global institutions or companies with the resources to carry out the digitisation and profit from it financially and socially. 

One should look to the entrenchment of the gains made in the last decade, taken together with correcting the gaps identified, whether infrastructural, environmental or incidental. Recognising that the current wave of AI activity has been formed on the basis of an NLP attractor, it is essential to find ways of institutionalising the NLP ecosystem in Kenya to ensure the independence and robustness of Kenya's technological development.

This calls for carefully considering governance and regulation of the country's AI and data ecosystem. In this developmental phase of technology, there is a need to develop flexible regulations whose aim is not to bureaucratise, stifle or gatekeep the industry but to guide it into the possibility of full flourishing. This has been discussed in the previous section, where the data protection act and the proposed AI and Robotics Bill have been seen as tentative steps taken by the Government of Kenya in this direction. There is an opportunity here for the Government to involve stakeholders more organically in ensuring robust legal infrastructure for the growth and takeoff of NLP and AI in the country.

\subsection{Research Opportunities: A road map to the Future}

\subsubsection{IKS and NLP in Climate Action}
The effectiveness of climate action depends on the extent to which available knowledge systems have been considered in the climate change discourse and their inclusivity in addressing the negative impacts of climate change. A fertile ground for revolutionising climate action strategies lies at the crossroads of Indigenous Knowledge Systems (IKS) and Natural Language Processing (NLP). In Kenya, Indigenous communities, stewards of their lands, offer a wealth of ecological wisdom shaped by centuries of sustainable living and intimate interactions with nature. Their practices, from agroforestry to water conservation, are a testament to a profound understanding of biodiversity and ecosystem management—key to combating climate change. However, a significant hurdle remains the need for datasets capturing these rich IKS worldviews, which hinders the potential of NLP applications. This data gap limits NLP technologies' ability to fully comprehend and utilise Indigenous languages and knowledge, excluding these vital perspectives from global climate action dialogues. In regions where indigenous knowledge has been integrated into technology, for instance, in the Amazon \cite{stejskal2022sources}, NLP has been used to document and translate local knowledge on medicinal plants, directly influencing sustainable practices and conservation efforts. Similarly, in Australia, Aboriginal fire management techniques, once overlooked, are now being studied and modelled through NLP \cite{varma2024machine}, providing insights into natural disaster management that are both ancient and innovative.
By bridging these gaps, particularly in Kenya, through dedicated research into creating comprehensive datasets of IKS, NLP can unlock a treasure trove of environmental strategies. Fusing traditional wisdom and modern technology can enrich our approaches to climate challenges and ensure a more inclusive and effective global response.

\section{Conclusions}
This paper underscores the importance of a concerted, long-term strategy for NLP in Kenya. This strategy includes expanding data resources, developing context-aware models, and engaging with Kenya's many languages' cultural and linguistic dimensions.
Our findings reveal significant gaps in data resources, language models, and the adaptation of existing NLP technologies to the unique linguistic contexts of Kenyan languages. Finally, we offer targeted recommendations for addressing these gaps, aiming to foster the development of more inclusive, accessible, and contextually relevant NLP solutions in Kenya.
To address these issues, we propose the following recommendations: a) a concerted effort should be made to expand the digital footprint of Kenyan languages through collaborative data collection and the creation of language-specific datasets. b) investments in local research and infrastructure are essential to empower Kenyan NLP Researchers and ensure that innovations are locally driven and sustainable. c) Integrating Indigenous knowledge systems into NLP workflows can provide critical linguistic and cultural context, enhancing model performance and ensuring the development of more accurate, context-aware, and culturally sensitive language technologies.

\section{Limitation}
Our work is derived from a literature survey on NLP research and publications about Kenyan languages. While we have covered many key publications, this review is not exhaustive. Due to the scope of our review, other relevant works and advancements in the field may not have been included. Further studies should aim to capture a more comprehensive view of the evolving landscape of NLP in Kenya.